\documentclass[letterpaper, 10 pt, conference]{ieeeconf}  

\IEEEoverridecommandlockouts                              

\usepackage{graphics} 
\usepackage{epsfig} 
\usepackage{mathptmx} 
\usepackage{times} 
\usepackage{amsmath} 
\usepackage{amssymb}  

\usepackage{cite}
\usepackage{amsmath,amssymb,amsfonts}
\usepackage{graphicx}
\usepackage{textcomp}
\usepackage{xcolor}
\usepackage{upgreek}
\usepackage{makecell}
\usepackage{algorithm}
\usepackage{algpseudocode}
\usepackage{prettyref}
\usepackage{bm}
\usepackage{pdfpages}
\usepackage{multirow}
\usepackage{booktabs}
\usepackage{siunitx}
\usepackage{subfigure}
\usepackage[percent]{overpic}
\usepackage{hyperref}

\newrefformat{fig}{Fig.~\ref{#1}}
\newrefformat{sec}{Sec.~\ref{#1}}
\newrefformat{tab}{Tab.~\ref{#1}}
\newrefformat{algo}{Algorithm.~\ref{#1}}
\newrefformat{eq}{Eq.~\ref{#1}}

\title{\LARGE \bf
Simultaneous Synchronization and Calibration for Wide-baseline Stereo Event Cameras
}
\author{
Wanli Xing$^{1}$, Shijie Lin$^{1}$, Guangze Zheng$^{1}$, Yanjun Du$^{2}$ and Jia Pan$^{1}$$^{\dagger}$
\thanks{
$^{1}$Wanli Xing, Shijie Lin, Guangze Zheng and Jia Pan are with the Department of Computer Science, The University of Hong Kong, Hong Kong SAR {\tt\footnotesize wlxing@connect.hku.hk}
\endgraf
$^{2}$Yanjun Du is with the Department of Mechanical and Automation Engineering, The Chinese University of Hong Kong, Hong Kong SAR
\endgraf
$^\dagger$Corresponding author
}
}

\begin{document}

\maketitle
\thispagestyle{empty}
\pagestyle{empty}

\begin{abstract}

Event-based cameras are increasingly utilized in various applications, owing to their high temporal resolution and low power consumption. However, a fundamental challenge arises when deploying multiple such cameras: they operate on independent time systems, leading to temporal misalignment. This misalignment can significantly degrade performance in downstream applications. Traditional solutions, which often rely on hardware-based synchronization, face limitations in compatibility and are impractical for long-distance setups. To address these challenges, we propose a novel algorithm that exploits the motion of objects in the shared field of view to achieve millisecond-level synchronization among multiple event-based cameras. Our method also concurrently estimates extrinsic parameters. We validate our approach in both simulated and real-world indoor/outdoor scenarios, demonstrating successful synchronization and accurate extrinsic parameters estimation. 

Code: \href{https://github.com/FORREST1901/ssac}{\textcolor[rgb]{0,0,0.9}{https://github.com/FORREST1901/ssac}}

\end{abstract}

\section{Introduction}
Event-based cameras have emerged as a groundbreaking technological development, inspired by the biological retina's mode of sensing \cite{reinbacher2016real}. In contrast to conventional frame-based cameras, which capture visual scenes at predetermined intervals, event-based cameras generate asynchronous events. These events are triggered only when the logarithmic intensity change of an individual pixel surpasses a pre-defined threshold \cite{gallego2020event}. This unique property endows event-based cameras with several compelling advantages: ultra-high temporal resolution, superior dynamic range, and minimal power consumption \cite{tedaldi2016feature}. Owing to their unparalleled capabilities, event-based cameras have proven exceptionally effective in niche applications, such as 3D human pose estimation \cite{zou2021eventhpe,calabrese2019dhp19} and 3D fluid flow reconstruction \cite{wang2020stereo}. In prospective domains like building information modeling \cite{azhar2012building} and wildlife monitoring\cite{ling2019behavioural}—where the primary focus is to capture fleeting dynamic events within predominantly static settings, the specialized attributes of event-based cameras position them as highly promising alternatives to conventional frame-based cameras. Specifically, their capability to selectively record swift changes in the environment suggests a strong potential for superior performance in these sectors.

\begin{figure}[t]
	\centering
	\subfigure[]{
		\begin{minipage}{\linewidth} 
            \centering
            \includegraphics[width=0.8\textwidth]{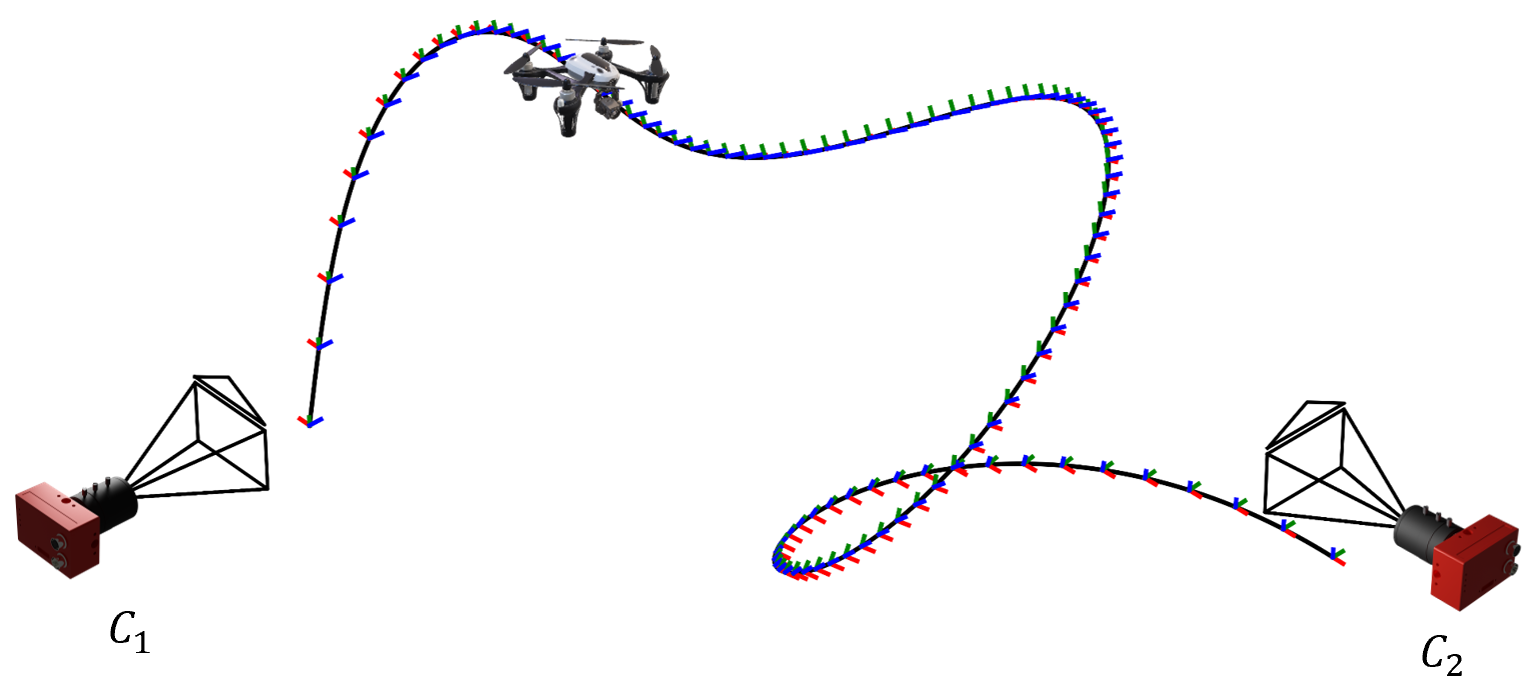} \\
            \vspace{-4mm}
		\end{minipage}}
	\subfigure[]{
		\begin{minipage}{\linewidth}
            \centering
            \includegraphics[width=0.8\textwidth]{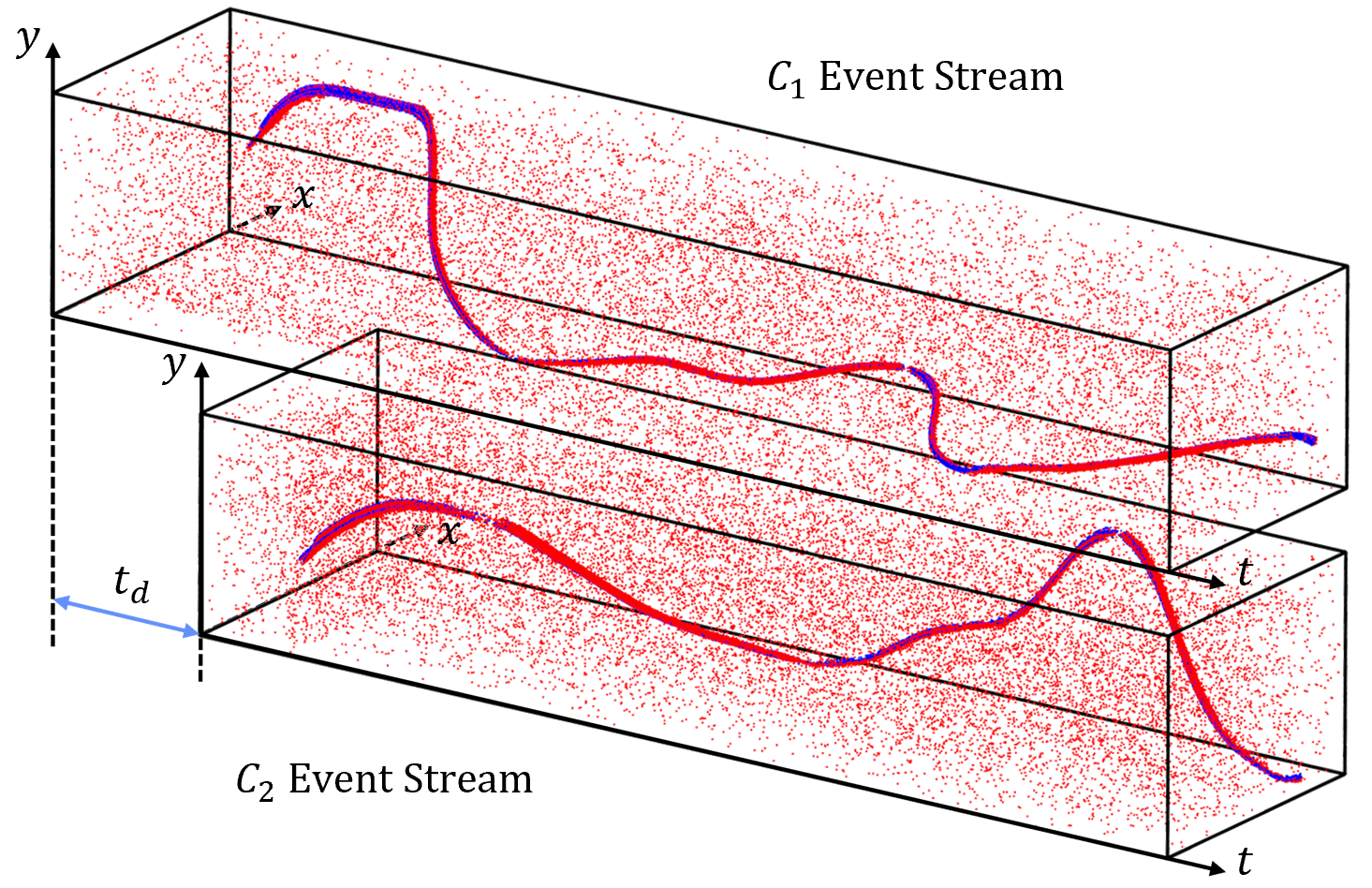} \\
            \vspace{-4mm}
		\end{minipage}}
	\caption{(a) Illustration of dual-perspective event capture: Two event-based cameras record events of a moving object from divergent viewpoints. The object's orientation along its trajectory is denoted by the local frame, represented by \textcolor[rgb]{0,  0,  0.9}{blue}-\textcolor[rgb]{0.9,  0,  0}{red}-\textcolor[rgb]{0,  0.7,  0}{green} axes. (b) Temporal misalignment: The event streams from the two cameras are asynchronously offset by a temporal discrepancy \(t_d\). The curves feature \textcolor[rgb]{0,  0,  0.9}{blue} and \textcolor[rgb]{0.9,  0,  0}{red} points, signifying the positive and negative events generated during the object's motion, whereas the scattered points illustrate the background noise.} 
 \label{fig:cam_and_trj}
\end{figure}

However, employing multiple event-based cameras within the same setting introduces intricate challenges related to temporal synchronization and extrinsic parameter calibration. Conventionally, the internal clock of an event-based camera is synchronized to a connected computational device. As a result, when two such cameras are not coherently synchronized, their time systems function independently. The ensuing time offset between these cameras hinges on the accuracy of synchronization between their associated computing devices, which can fluctuate by hundreds of milliseconds or even seconds. Given the ultra-high temporal resolution inherent to event-based cameras, the margin for error in their synchronization is significantly smaller than that for traditional frame-based cameras. This synchronization challenge is further compounded in wide-baseline configurations, where the spatial separation between cameras amplifies both synchronization and calibration issues. Most extant solutions rely on hardware-level synchronization techniques, such as specialized synchronization cables or external timing signals like GPS \cite{gehrig2021dsec,perot2020learning,zhu2018realtime}. These hardware-based synchronization methods encounter several constraints, such as limited compatibility across diverse event camera models and the challenges associated with implementing extensive cable configurations over long distances.

To address existing limitations and challenges, we present a novel, software-centric method for temporally synchronizing event-based cameras. As illustrated in \prettyref{fig:cam_and_trj}, consider two event-based cameras, $C_1$ and $C_2$, capturing an object within their overlapping fields of view. These cameras generate event streams that are misaligned by a temporal offset, $t_d$. Our approach exploits epipolar geometry to correlate these asynchronous event streams. Specifically, we employ the epipolar geometric constraints to formulate an optimization problem aimed at precisely determining the temporal offset $t_d$. This software-based technique effectively eliminates the requirement for dedicated hardware, thus expanding the range of potential applications for multi-camera event-based systems—particularly in setups with wide baselines or substantial inter-camera separation. Importantly, our methodology also permits the simultaneous estimation of extrinsic parameters. The robustness of our proposed synchronization approach is empirically substantiated through extensive evaluations, including both simulation-based benchmarks and real-world experimental trials.

In summary, the primary contributions of this paper are as follows:
\begin{itemize}
\item We propose a novel approach for temporal synchronization of event-based cameras in wide-baseline settings, thereby obviating the necessity for dedicated hardware components.
\item Our methodology not only synchronizes the cameras but also concurrently estimates extrinsic parameters.
\item Extensive experiments, both in simulation and real-world environments, validate the efficacy and applicability of the proposed framework.
\end{itemize}

\section{Related Work}

The reliable synchronization in stereo vision generally falls into two categories: hardware-based and data-driven methodologies.

\subsection{Hardware and Data-Driven Synchronization for Event-Based Cameras}

Predominantly, manufacturers of event-based cameras incorporate synchronization interfaces to facilitate multi-camera setups. These interfaces usually permit synchronization through the cameras' internal pulse signals \cite{zhu2018multivehicle} or external triggers such as GPS signals \cite{gehrig2021dsec}.

\cite{censi2014low} employed a rudimentary synchronization technique that calculates the event rate in two distinct event streams by enumerating events over fixed time intervals. Although this method proves adequate for scenarios requiring modest synchronization accuracy, it falls short in applications demanding higher precision. Another approach by \cite{perot2020learning} leverages zero-normalized cross-correlation (ZNCC) to synchronize two event-based cameras. This method relies on one-dimensional statistics derived from both event streams and standard images, subsequently utilizing these for ZNCC computation. Both methods presume that the event-based cameras in question are capturing similar scenes and exhibit consistent motion—a condition that is often untenable for statically positioned cameras capturing divergent scenes.

\subsection{Synchronization Strategies for Frame-Based Cameras}

Given the ubiquity of frame-based cameras, a plethora of synchronization algorithms have been devised to accommodate them. These include methods that exploit common audio cues \cite{shrstha2007synchronization} or detect flashes in multiple video feeds \cite{shrestha2006synchronization}. Another innovative approach involves the use of silhouettes for camera calibration. In this context, a human subject is placed against a monochromatic backdrop and adopts varying poses while being captured from multiple viewpoints. The extracted silhouette aids in action reconstruction \cite{bottino2001silhouette,natsume2019siclope}, and can be further employed to locate epipolar lines, thus facilitating the calculation of extrinsic camera parameters \cite{boyer2006using}. \cite{sinha2010camera} expanded this silhouette-based approach to accommodate unsynchronized video streams in monochromatic settings, achieving synchronization accuracy suitable for human motion tracking.

\section{Problem Formulation}

Accurate temporal synchronization among vision sensors—such as event-based and frame-based cameras—is crucial for a high-fidelity stereo vision system. Misaligned temporal data can introduce substantial errors in tasks like scene depth estimation and object motion analysis. Hence, robust solutions for sensor synchronization are imperative for maintaining the integrity of stereo vision applications.

\subsection{Frame-Based Camera Synchronization}

\begin{figure}[t]
	\centering
	\subfigure[Synchronization of frame-based cameras]{
		\begin{minipage}{\linewidth} 
            \centering
            \label{fig:rgb_def}
            \includegraphics[width=\textwidth]{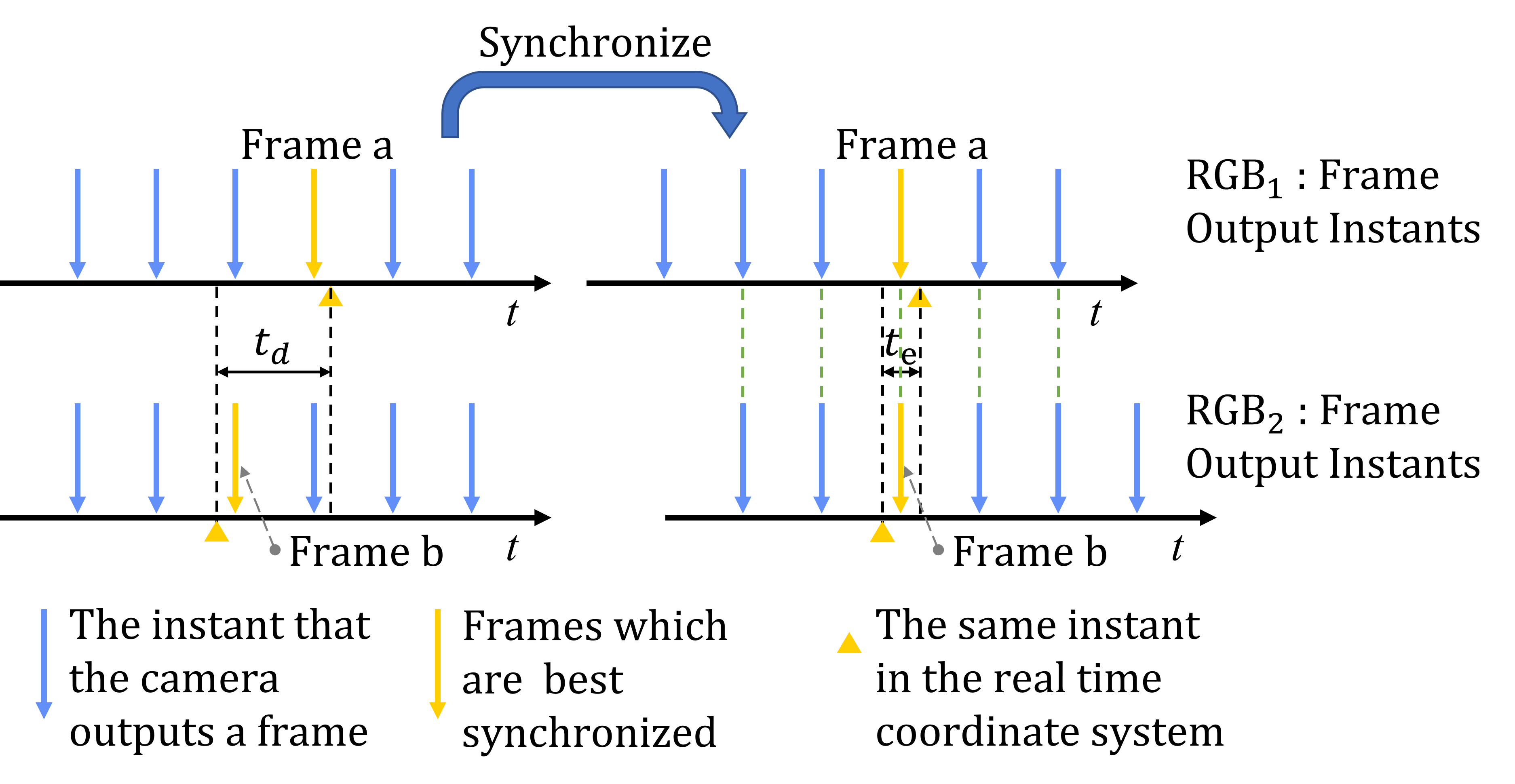} \\
		\end{minipage}}
	\subfigure[Synchronization of event-based cameras]{
		\begin{minipage}{\linewidth}
            \centering
            \label{fig:eve_def}
            \includegraphics[width=\textwidth]{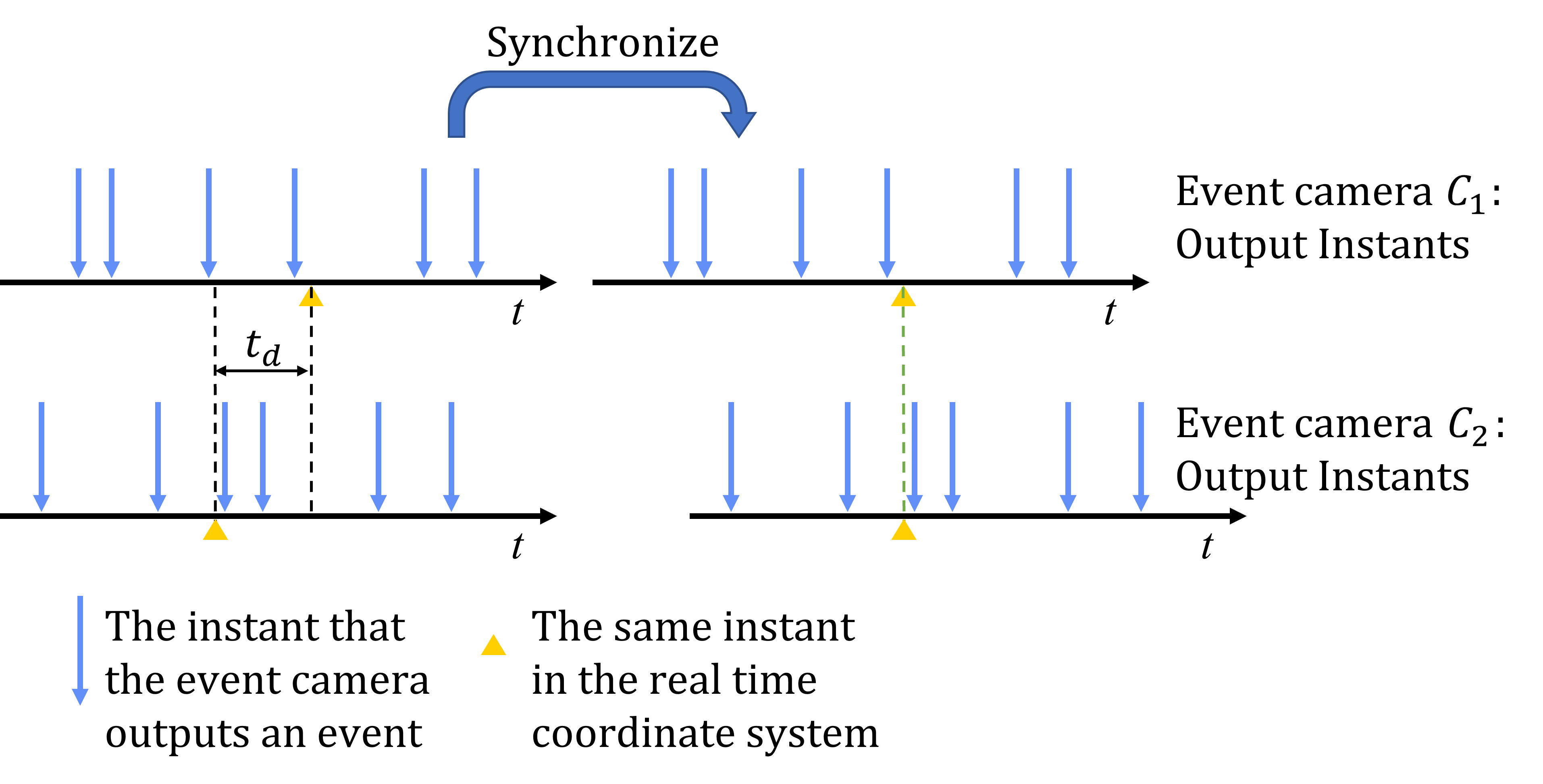} \\
		\end{minipage}}
	\caption{
 (a) Illustration of frame-based camera synchronization. Each arrow denotes the moment an image is captured. Given the fixed frame rate, inter-arrow distances remain consistent. Yellow triangles indicate corresponding real-time instants, while the yellow arrows highlight the frames best aligned for synchronization. \(t_d\) represents the ground-truth temporal synchronization offset, and \(t_e\) is the residual offset post-synchronization. (b) Illustration of event-based camera synchronization. Each arrow signifies the moment an event is triggered. Unlike frame-based cameras, event-based cameras generate events at varying intervals, dependent on intensity changes. Yellow triangles represent corresponding real-time instants. \(t_d\) denotes the ground-truth temporal synchronization offset.} 
 \label{fig:problem_def}
\end{figure}

In a stereo frame-based camera system without hardware-level synchronization, each camera functions independently, capturing images at a constant frequency \(f_c\). As depicted in  \prettyref{fig:rgb_def}, this inherent asynchrony calls for a temporal alignment process to synchronize the video streams between the two cameras. The primary objective of this alignment is to identify corresponding frames from both streams, such that frame \(a\) from camera RGB\(_1\) temporally aligns with frame \(b\) from camera RGB\(_2\).

It is noteworthy that the temporal markers of the captured frames in both RGB\(_1\) and RGB\(_2\) may not be perfectly aligned. Consequently, even after establishing the most accurate frame-to-frame correspondence, the residual temporal offset could be as large as \( \frac{1}{2f_c} \).

\subsection{Event-Based Camera Synchronization}
Synchronizing stereo event-based camera systems is challenging. This is due to each camera's independent internal control system, which generates timestamps for events that can differ even if occurring simultaneously. Like its frame-based counterpart, an event-based stereo camera system also has an inherent but unknown time offset between its event streams.

As illustrated in \prettyref{fig:eve_def}, unsynchronized event-based cameras yield event streams with timestamps that differ by a constant \( t_d \). Mathematically, this relationship can be expressed as:

\begin{equation}
    t_{C1} = t_{C2} + t_d.
    \label{eq:td definition}
\end{equation}

In this study, we aim to optimally estimate \( t_d \) by analyzing the motion of a drone within the overlapping field of view of two event-based cameras.

\section{Methodology}
We introduce a novel approach for outdoor synchronization of distant event-based cameras without needing specialized calibration boards. Our method utilizes a moving object, like a drone or bird, within the cameras' shared field of view. By estimating the object's geometric center in each camera view and leveraging epipolar geometry, we can compute the time offset between the cameras. This approach is applicable even when the cameras' extrinsic parameters are unknown and can simultaneously estimate these parameters during synchronization.

Initially, we elucidate the methodology for acquiring the motion trajectories of the object as captured by each event-based camera, as detailed in \prettyref{sec:The Object Motion}. Subsequently, in \prettyref{sec:Synchronization with Known Camera Extrinsic Parameters}, we elaborate on the synchronization procedure under the premise of known extrinsic parameters. Lastly, \prettyref{sec:Synchronization without Known Camera Extrinsic Parameters} delineates the approach to synchronization when the extrinsic parameters remain undetermined.

\subsection{The Object Motion}
\label{sec:The Object Motion}

We exploit the principle that the geometric center of an object's event stream can serve as an accurate representation of the object's position under specific conditions. These conditions include either a small angular footprint of the object in the camera's view or symmetrical object shapes. As illustrated in \prettyref{fig:cam_and_trj}, the distance of drones or birds from the event-based cameras in outdoor settings often renders the diameter of their corresponding event stream to a mere few pixels. This allows us to use the geometric center as a reliable approximation of the object's position.

\begin{figure}[t]
	\centering
	\subfigure[Original event stream]{
		\begin{minipage}{\linewidth} 
            \includegraphics[width=\textwidth]{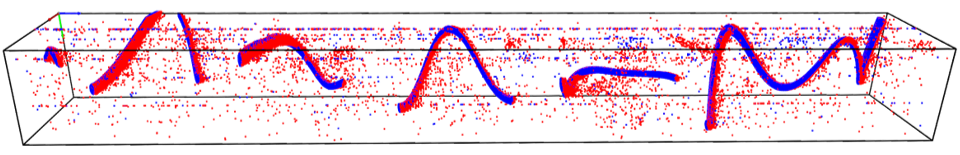} \\
            \vspace{-4mm}
		\end{minipage}}
	\subfigure[Event stream after denoising]{
		\begin{minipage}{\linewidth}
			\includegraphics[width=\textwidth]{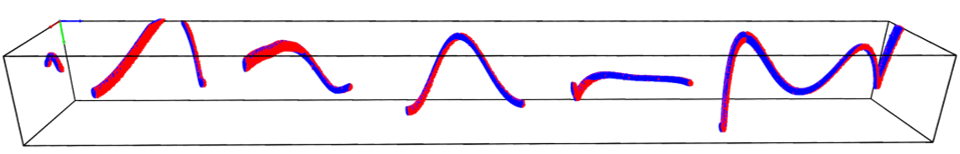} \\
			\vspace{-4mm}
		\end{minipage}}
	\subfigure[2D trajectory in $x$-$y$-$t$ coordinate]{
		\begin{minipage}{\linewidth}
			\includegraphics[width=\textwidth]{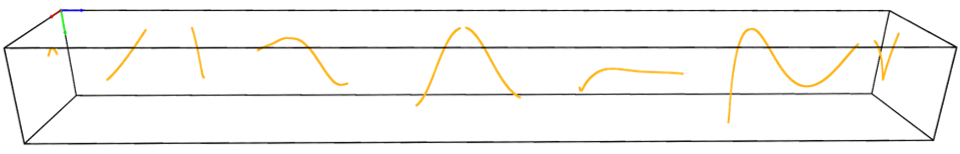} \\
			\vspace{-4mm}
		\end{minipage}}
	\caption{Event stream preprocessing workflow for object motion extraction. (a) Raw event stream; (b) Event stream post-statistical outlier removal; (c) Refined 2D pose at each temporal instance.}
 	\label{fig:pre_process}

\end{figure}

The event-based cameras naturally generate an event stream when exposed to variations in background illumination or brightness. Noise from the transistor circuitry can also trigger events. To isolate events related to object motion, we employ a filtering process in the $x$-$y$-$t$ space, eliminating events with insufficient nearby events within a given radius. Additionally, we correct for optical distortion in the $x$ and $y$ coordinates, for which the distortion and intrinsic parameters are obtained through individual camera calibration.

For a specified temporal instance $t_0$, we determine the object's $x$-$y$ position on the image plane by averaging the 2D coordinates of all triggered events in a brief time window surrounding $t_0$. This excludes events under the following abnormal conditions:
\begin{itemize}
\item Events occurring on the image boundary imply the object is either entering or leaving the field of view; these events are not considered for synchronization.
\item When the event count within the time interval is less than a specified threshold $N_{min}$, it suggests slow object motion. Consequently, the average coordinates become less reliable and are thus disregarded for synchronization.
\end{itemize}

Upon completing these preparatory steps (as delineated in \prettyref{fig:pre_process}), we acquire the requisite object motion data to proceed with the subsequent synchronization algorithm.

\subsection{Synchronization when Camera Extrinsic Parameters are Known}
\label{sec:Synchronization with Known Camera Extrinsic Parameters}

In the following discussion, we initially focus on estimating the temporal offset between two event-based cameras when their extrinsic parameters are given a priori. Subsequently, in \prettyref{sec:Synchronization without Known Camera Extrinsic Parameters}, we delve into the methodologies for determining these extrinsic parameters themselves.

As illustrated in \prettyref{fig:cam_and_trj}, the spatial trajectory of a moving object in three-dimensional space yields distinct two-dimensional trajectories on the image planes of individual event-based cameras. These trajectories are essentially projections of the 3D path viewed from differing orientations. Assuming two such cameras, denoted as \(C_1\) and \(C_2\), are perfectly synchronized, then at any given time \(t\), the points \(\mathbf{p}_1(t)\) and \(\mathbf{p}_2(t)\) on these projected 2D paths should satisfy the epipolar geometric constraints \cite{andrew2001multiple}:

\begin{equation}
\mathbf{p}_2^T(t)\mathbf{F}\mathbf{p}_1(t)=0,
    \label{eq:epipolar}
\end{equation}
where \(\mathbf{F}\) represents the fundamental matrix, defined as \(\mathbf{F}=\mathbf{K}_2^{-T}[\mathbf{t}]_{\times}\mathbf{R}\mathbf{K}_1^{-1}\). Here, \(\mathbf{K}_1\) and \(\mathbf{K}_2\) are the intrinsic parameter matrices for \(C_1\) and \(C_2\), respectively, while \(\mathbf{R}\) and \(\mathbf{t}\) denote their relative rotation and translation.

In a more generalized scenario where a temporal discrepancy \(t_d\) exists between the timestamps of \(C_1\) and \(C_2\), the projected points \(\mathbf{p}_1(t)\) and \(\mathbf{p}_2(t)\) are governed by a modified constraint:

\begin{equation}
\mathbf{p}_2^{T}(t+t_d)\mathbf{F}\mathbf{p}_1(t)=0.
    \label{eq:epipolar_td}
\end{equation}

\begin{figure}[t]
    \centering
    \hspace{-4mm}
    \subfigure[$C_1$ Event Image]{
        \begin{overpic}[width=0.239\textwidth]{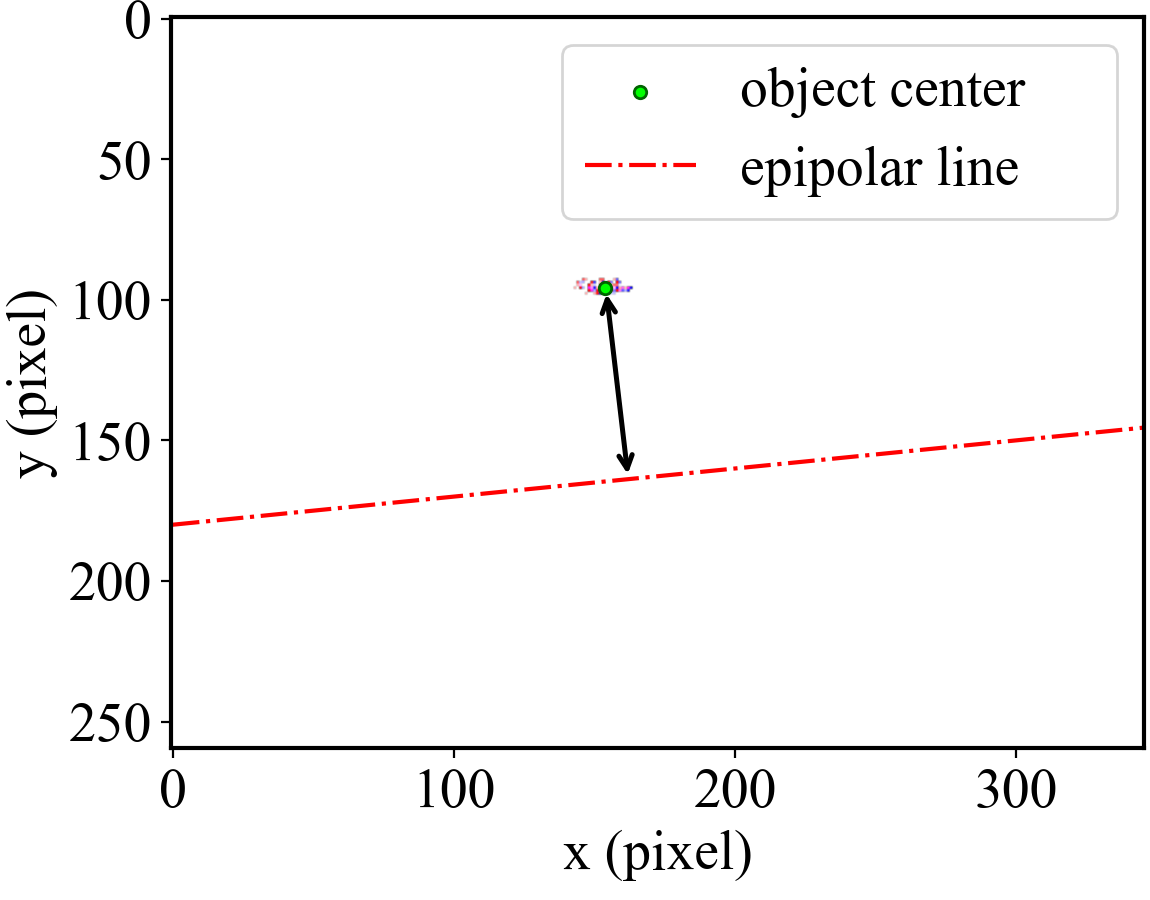}
            \put(91.7,70){\tiny $\mathbf{p}_1$}
            \put(91.7,62.5){\tiny $\mathbf{l}_1$}
            \put(58,44){\tiny $d=d(\mathbf{p}_1,\mathbf{l}_1)$}
        \end{overpic}
    }
    \hspace{-3mm}
    \subfigure[$C_2$ Event Image]{
        \begin{overpic}[width=0.239\textwidth]{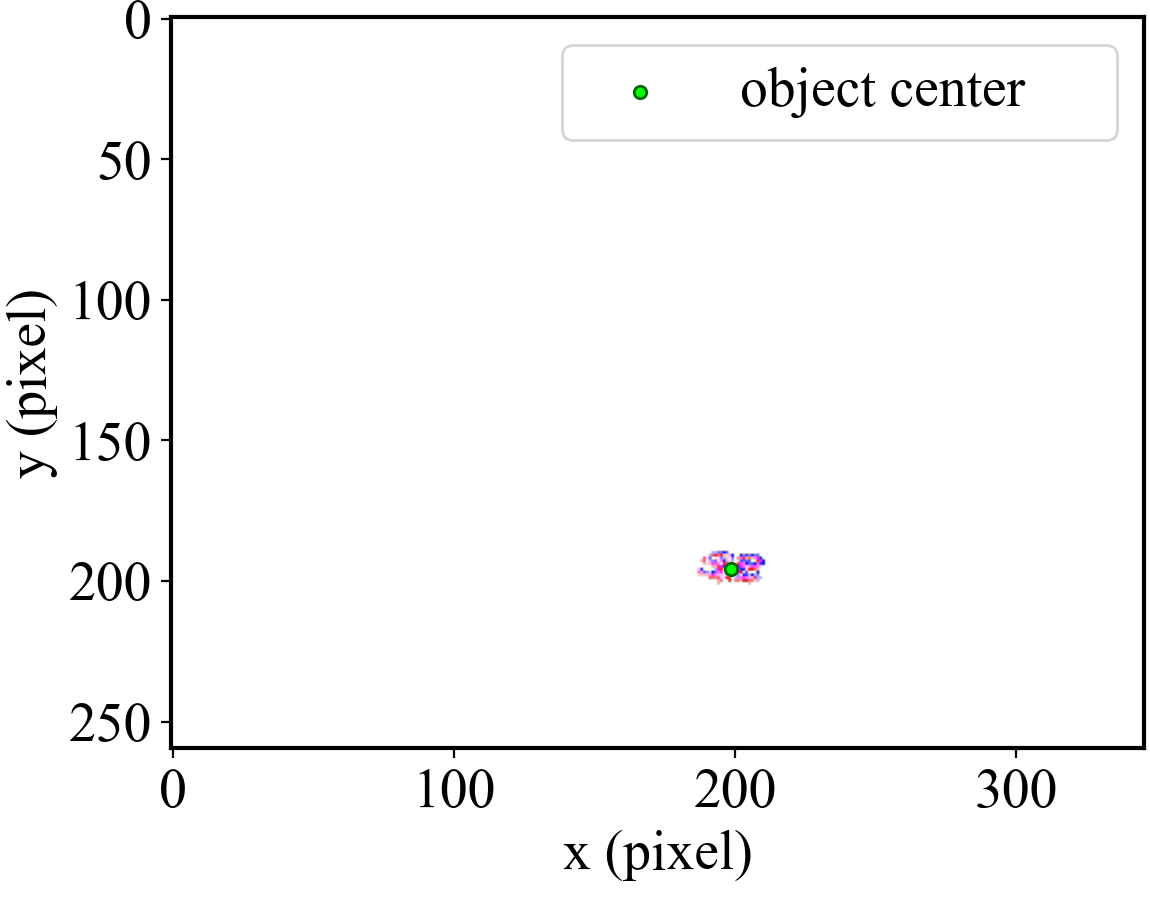}
            \put(91.7,70){\tiny $\mathbf{p}_2$}
        \end{overpic}
    }
    \caption{
        Illustration of the distance \( d \) between the object center \( \mathbf{p}_1 \) and the epipolar line \( \mathbf{l}_1 \) in the image plane of camera \( C_1 \). (a) depicts the event-based image from \( C_1 \), highlighting both the object center \( \mathbf{p}_1 \) and the epipolar line \( \mathbf{l}_1 \). (b) displays the event-based image from \( C_2 \), along with the object center \( \mathbf{p}_2 \). The pixels colored in red and blue are indicative of events.
    }
    \label{fig:epipolar}
\end{figure}

To evaluate the extent of constraint violation when \(t_d\) is suboptimal, we introduce a metric that quantifies the distance \(d(\mathbf{p}_1(t),\mathbf{l}_1(t))\) between the corresponding epipolar line \(\mathbf{l}_1(t)\) and the point \(\mathbf{p}_1(t)\) (see \prettyref{fig:epipolar}). The epipolar line \( \mathbf{l}_1(t) \) is formally given by: \(\mathbf{l}_1(t_i) = \mathbf{F} \mathbf{p}_2(t_i + t_d)\). This line is determined by the relative positions of camera \( C_1 \) and \( C_2 \), as well as the position of \( \mathbf{p}_2 \). By aggregating this metric across a range of time instances \(t_i\), we derive an average quality measurement for \(t_d\):

\begin{equation}
    d_{\text{avg}} = \frac{1}{N}\sum_{i=1}^{N} d\left(\mathbf{p}_1(t_i),\mathbf{l}_1(t_i)\right).
    \label{eq:d_avg}
\end{equation}

\begin{figure}[t]
    \centering
    \hspace{-5mm}
    \subfigure[$d_{\text{avg}}$ from -1 s to 1 s]{
        \begin{overpic}[width=0.239\textwidth]{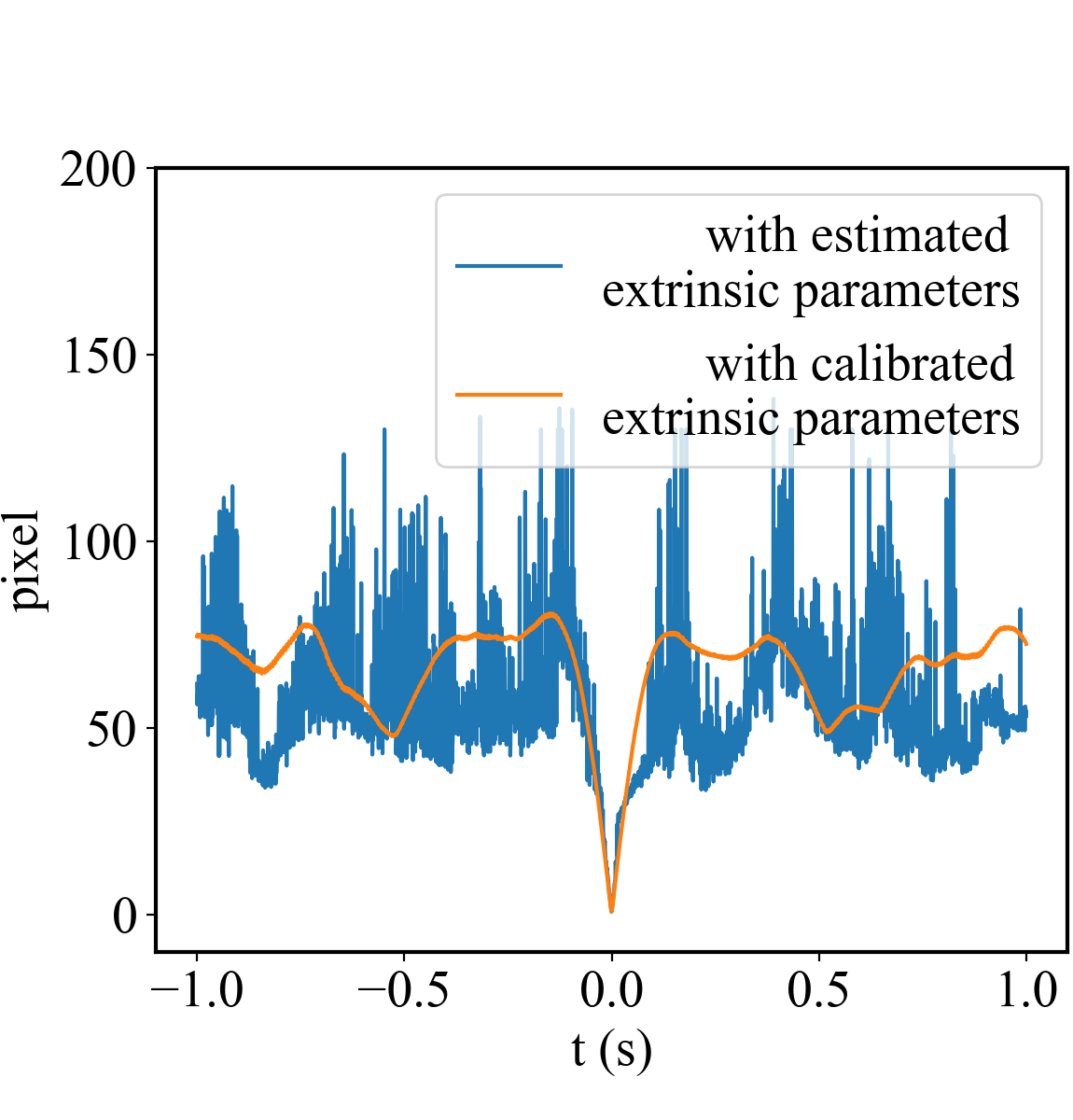}
            \put(54,78){\tiny $d_{\text{avg}}$}
            \put(54,66.5){\tiny $d_{\text{avg}}$}
        \end{overpic}
    }
    \subfigure[$d_{\text{avg}}$ from -3 ms to 3 ms]{
        \begin{overpic}[width=0.239\textwidth]{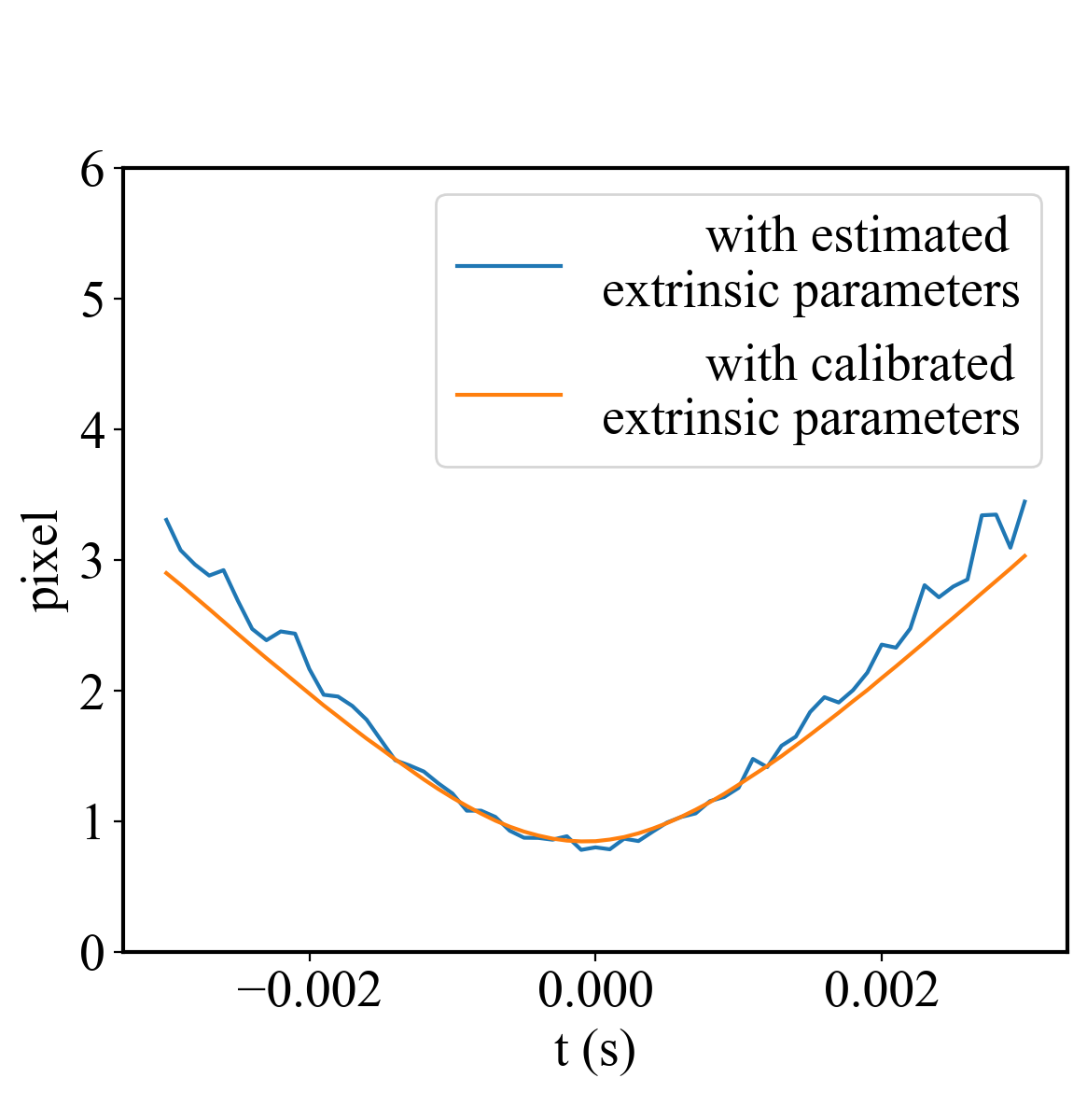}
            \put(54,78){\tiny $d_{\text{avg}}$}
            \put(54,66.5){\tiny $d_{\text{avg}}$}        
        \end{overpic}
    }
    \caption{
        Visualization of the average distance \( d_{\text{avg}} \) for well-synchronized (the ground-truth \( t_d = 0 \)) event streams. (a) plots \( d_{\text{avg}} \) over a range of -1 s to 1 s, while (b) zooms in to the interval between -3 ms and 3 ms. The minimum \( d_{\text{avg}} \) value is observed to occur near \( t_d = 0 \).
    }
    \label{fig:avg_dist}
\end{figure}

The algorithm for identifying the optimal \( t_d \) within the potential search interval \([t_{b}, t_{e}]\) is detailed in \prettyref{algo:sychronization_with_known_extrinsic}. For each \( t_d \), we sample \( N \) time points \( t_i \), compute the average distance \( d_{\text{avg}} \) between each point \( \mathbf{p}_1(t_i) \) and the epipolar line \( \mathbf{l}_1(t_i) \), and identify the \( t_d \) that minimizes \( d_{\text{avg}} \). \prettyref{fig:avg_dist} shows that the minimal \( d_{\text{avg}} \) aligns closely with the true time offset.

\begin{algorithm}[t]
	\begin{algorithmic}[1]
		\Require  
		The event streams from two event-based cameras
		\Ensure  
		The time offset between two event-based cameras
		
		\For{$t_d\in [t_{b}, t_{e}]$}
		\State Take $t_d$ as time offset;
            \State Sample $N$ distinct time points, $t_i$, within the overlapping duration;
    		\For{$i=1$ to $N$}
    		\State $\mathbf{l}_1(t_i)=\mathbf{F}\mathbf{p}_2(t_i+t_d)$;
    		\State $d_i=d(\mathbf{l}_1(t_i),\mathbf{p}_1(t_i))$;
    		\EndFor
		\State $d_{avg}=\frac{1}{N}\sum_{i=1}^{N}d_i$;
		\EndFor
		\State Find the minimal $d_{avg}$ and the corresponding $t_d^*$.
	\end{algorithmic}
	\caption{Synchronization when  Camera Extrinsic Parameters are Known}
	\label{algo:sychronization_with_known_extrinsic} 
\end{algorithm}

\subsection{Synchronization with Unknown Camera Extrinsic Parameters}
\label{sec:Synchronization without Known Camera Extrinsic Parameters}

In scenarios where the extrinsic parameters—namely the rotation matrix \( \mathbf{R} \) and translation vector \( \mathbf{t} \)—between two event-based cameras are unspecified, the fundamental matrix \( \mathbf{F} \) must be estimated to establish the epipolar constraint. Given a temporal offset \( t_d \), the optimal value of which satisfies:
\begin{equation}
    \mathbf{p}_2^{T}(t + t_d)\mathbf{F}(t_d)\mathbf{p}_1(t) = 0.
    \label{eq:epipolar_td_estimated_F}
\end{equation}

This equation introduces three linear constraints on the elements of \( \mathbf{F}(t_d) \). Sampling \( N \) moments \( t_i \) where the object is visible to both cameras yields \( N \) pairs of points \( \mathbf{p}_1(t_i) \) and \( \mathbf{p}_2(t_i + t_d) \). These pairs form a linear system, solvable using robust methods such as the Least Median of Squares (LMedS) \cite{Rousseeuw1984}, to estimate \( \mathbf{F}(t_d) \). This estimated \( \mathbf{F}(t_d) \) is then utilized as per \prettyref{sec:Synchronization with Known Camera Extrinsic Parameters} to assess \( t_d \) and pinpoint its optimal value. Details are provided in \prettyref{algo:sychronization_without_known_extrinsic}.

As an ancillary outcome, decomposing \( \mathbf{F} \) at the optimal \( t_d^* \) yields the rotation \( \mathbf{R} \) and translation \( \mathbf{t} \) between the cameras. However, due to the scale ambiguity of \( \mathbf{F} \), \( \mathbf{t} \) indicates only the relative orientation, not the absolute distance, between the two coordinate systems.

\begin{algorithm}[t]
	\begin{algorithmic}[1]
		\Require  
		The event streams from two event-based cameras
		\Ensure  
		The time offset between two event-based cameras and the extrinsic parameters
		
		\For{$t_d\in [t_{b}, t_{e}]$}
		\State Take $t_d$ as time offset;
            \State Sample $N$ distinct time points, $t_i$, within the overlapping duration;
		\State Estimate the fundamental matrix $\mathbf{F}(t_d)$;
    		\For{$i=1$ to $N$}
    		\State $\mathbf{l}_1(t_i)=\mathbf{F}(t_d)\mathbf{p}_2(t_i+t_d)$;
    		\State $d_i=d(\mathbf{l}_1(t_i),\mathbf{p}_1(t_i))$;
    		\EndFor
		\State $d_{avg}=\frac{1}{N}\sum_{i=1}^{N}d_i$;
		\EndFor
		\State Find the minimal $d_{avg}$ and the corresponding $t_d^*$;
		\State Decompose the fundamental matrix $\mathbf F(t_d^*)$ to get extrinsic parameters $\mathbf R$ and $\mathbf t$.
	\end{algorithmic}
	\caption{Synchronization when  Camera Extrinsic Parameters are Unknown}
	\label{algo:sychronization_without_known_extrinsic} 
\end{algorithm}

\section{Experimental Results}

This section presents a comprehensive evaluation of the proposed algorithm for temporal synchronization and camera extrinsic parameter calibration. We begin by assessing the algorithm's performance in a simulated environment, then validate it experimentally using a physical setup featuring two event-based cameras and a 3D-printed miniature drone. Finally, the algorithm is further tested in an outdoor setting with a real drone, comparing calculated and ground truth trajectories.

\subsection{Simulation Experiments}

We employ ESIM \cite{Rebecq18corl}, a open-source event-based camera simulator, to generate synthetic event data. The simulated setup comprises two stationary event-based cameras and a moving drone, all of which have a resolution of \( 346 \times 260 \), akin to the DAVIS346 model.

To scrutinize the algorithm's robustness with respect to varying camera extrinsic parameters, we systematically alter these parameters in our simulations. Specifically, the angle between the cameras' lines of sight is varied from \( 30^\circ \) to \( 150^\circ \) in increments of \( 30^\circ \). Additionally, to assess the impact of event stream duration on algorithmic performance, tests are conducted using streams of both 10 and 20 seconds. Throughout these tests, the drone follows a random spline trajectory as proposed in \cite{Rebecq18corl}. To systematically introduce the time offset, \( t_d \), we predefine its values to be 5 ms, 50 ms, or 500 ms and integrate this offset into the event timestamps of one of the data streams.

For each experimental condition, the proposed algorithm is run 10 times and the average results, along with their standard deviations, are summarized in \prettyref{tab:esim_10} and \prettyref{tab:esim_20}. Our time offset search is conducted within \( \pm 1000 \) ms of the ground truth value. To evaluate the accuracy of estimated extrinsic parameters, we compute the angular discrepancies between the estimated and true rotation and translation vectors.

We also compare the proposed algorithm's synchronization accuracy with that of the zero-normalized cross-correlation (ZNCC) method \cite{perot2020learning}. While ZNCC uses both the event stream and frame images to generate a one-dimensional signal, resulting in a minimum synchronization unit equal to the inter-frame interval (20 ms for a 50 Hz frame rate), our approach allows for finer granularity. For comparison, we designate the frame-based and event-based ZNCC methods as ZNCC\(_F\) and ZNCC\(_E\), respectively. Their performance metrics are also included in \prettyref{tab:esim_10} and \prettyref{tab:esim_20}.

\begin{table}[t]
\centering
\caption{10-Second Trajectory Simulation}
\label{tab:esim_10}
\scalebox{0.73}{
\begin{tabular}{|c|c|c|c|c|c|c|}
\hline
\multirow{2}{*}{Angle[°]} &
\multirow{2}{*}{$t_d$[ms]} &
\multicolumn{3}{c|}{$t_{d\ error}$[ms]} &
\multirow{2}{*}{$R_{error}$[°]} &
\multirow{2}{*}{$t_{error}$[°]}\\
\cline{3-5}
    &   & \textbf{ours} & ZNCC$_E$  \cite{perot2020learning}  & ZNCC$_F$  \cite{perot2020learning}  &    &   \multirow{2}{*}{} \\
\hline
\multirow{3}{*}{30} & 5     & \textbf{0.16 (0.21)} & 17.0 (11.7) & 21.0 (16.6) & 0.4 (0.33)  & 0.43 (0.45)  \\ \cline{2-7} 
                    & 50    & \textbf{0.21 (0.16)} & 17.0 (14.2) & 20.6 (17.9) & 0.41 (0.33)  & 0.37 (0.42)  \\ \cline{2-7} 
                    & 500   & \textbf{0.18 (0.17)} & 30.0 (38.2) & 51.9 (92.0) & 0.28 (0.2)  & 0.52 (0.57)  \\ \cline{2-7} 
\hline
\multirow{3}{*}{60} & 5     & \textbf{0.17 (0.22)} & 43.0 (35.4) & 99.6 (151.2) & 0.55 (0.44)  & 0.33 (0.33)  \\ \cline{2-7} 
                    & 50    & \textbf{0.2 (0.22)} & 47.0 (44.5) & 95.2 (138.3) & 0.72 (0.67)  & 0.44 (0.49)  \\ \cline{2-7} 
                    & 500   & \textbf{0.28 (0.24)} & 64.0 (80.8) & 95.5 (125.0) & 0.61 (0.36)  & 0.27 (0.15)  \\ \cline{2-7} 
\hline
\multirow{3}{*}{90} & 5     & \textbf{0.24 (0.14)} & 272.0 (346.0) & 346.4 (365.4) & 0.39 (0.15)  & 0.19 (0.15)  \\ \cline{2-7} 
                    & 50    & \textbf{0.16 (0.15)} & 221.0 (259.9) & 349.6 (367.2) & 0.47 (0.21)  & 0.24 (0.22)  \\ \cline{2-7} 
                    & 500   & \textbf{0.34 (0.25)} & 265.0 (260.2) & 387.9 (383.4) & 0.45 (0.12)  & 0.28 (0.19)  \\ \cline{2-7} 
\hline
\multirow{3}{*}{120} & 5    & \textbf{0.35 (0.25)} & 415.0 (362.6) & 616.6 (405.1) & 0.37 (0.26)  & 0.18 (0.16)  \\ \cline{2-7} 
                    & 50    & \textbf{0.22 (0.12)} & 410.0 (356.7) & 585.4 (388.7) & 0.53 (0.57)  & 0.29 (0.32)  \\ \cline{2-7} 
                    & 500   & \textbf{0.29 (0.22)} & 437.0 (368.4) & 505.3 (383.6) & 0.55 (0.31)  & 0.26 (0.16)  \\ \cline{2-7} 
\hline
\multirow{3}{*}{150} & 5    & \textbf{0.28 (0.21)} & 454.0 (248.7) & 700.0 (298.5) & 0.52 (0.29)  & 0.3 (0.15)  \\ \cline{2-7} 
                    & 50    & \textbf{0.49 (0.4)} & 407.0 (253.6) & 790.7 (231.4) & 0.6 (0.32)  & 0.33 (0.14)  \\ \cline{2-7} 
                    & 500   & \textbf{0.58 (0.39)} & 381.0 (140.2) & 687.7 (279.4) & 0.62 (0.36)  & 0.31 (0.24)  \\ \cline{2-7} 
\hline
\end{tabular}
}
\end{table}

\begin{table}[t]
\centering
\caption{20-Second Trajectory Simulation}
\label{tab:esim_20}
\scalebox{0.73}{
\begin{tabular}{|c|c|c|c|c|c|c|}
\hline
\multirow{2}{*}{Angle[°]} &
\multirow{2}{*}{$t_d$[ms]} &
\multicolumn{3}{c|}{$t_{d\ error}$[ms]} &
\multirow{2}{*}{$R_{error}$[°]} &
\multirow{2}{*}{$t_{error}$[°]}\\
\cline{3-5}
    &   & \textbf{Ours} & ZNCC$_{E}$ \cite{perot2020learning} & ZNCC$_{F}$ \cite{perot2020learning} &    &   \multirow{2}{*}{} \\
\hline
\multirow{3}{*}{30} & 5     & \textbf{0.13 (0.05)} & 15.0 (11.0) & 20.5 (13.2) & 0.16 (0.07)  & 0.19 (0.14)  \\ \cline{2-7} 
                    & 50    & \textbf{0.11 (0.05)} & 15.0 (12.0) & 20.5 (13.4) & 0.14 (0.05)  & 0.15 (0.14)  \\ \cline{2-7} 
                    & 500   & \textbf{0.13 (0.09)} & 28.0 (46.0) & 50.6 (89.4) & 0.18 (0.05)  & 0.2 (0.1)  \\ \cline{2-7} 
\hline
\multirow{3}{*}{60} & 5     & \textbf{0.13 (0.1)} & 143.0 (227.9) & 113.5 (204.9) & 0.25 (0.08)  & 0.18 (0.09)  \\ \cline{2-7} 
                    & 50    & \textbf{0.11 (0.07)} & 111.0 (221.6) & 115.4 (204.6) & 0.22 (0.05)  & 0.15 (0.11)  \\ \cline{2-7} 
                    & 500   & \textbf{0.11 (0.07)} & 176.0 (242.0) & 154.3 (218.0) & 0.26 (0.1)  & 0.18 (0.12)  \\ \cline{2-7} 
\hline
\multirow{3}{*}{90} & 5     & \textbf{0.36 (0.2)} & 543.0 (397.5) & 582.5 (360.7) & 0.28 (0.11)  & 0.15 (0.08)  \\ \cline{2-7} 
                    & 50    & \textbf{0.31 (0.13)} & 616.0 (368.5) & 583.2 (359.7) & 0.32 (0.12)  & 0.16 (0.09)  \\ \cline{2-7} 
                    & 500   & \textbf{0.21 (0.14)} & 590.0 (362.2) & 640.3 (324.7) & 0.29 (0.08)  & 0.17 (0.11)  \\ \cline{2-7} 
\hline
\multirow{3}{*}{120} & 5    & \textbf{0.24 (0.22)} & 716.0 (267.1) & 796.5 (180.3) & 0.3 (0.12)  & 0.18 (0.1)  \\ \cline{2-7} 
                    & 50    & \textbf{0.23 (0.17)} & 654.0 (319.2) & 817.0 (153.0) & 0.26 (0.09)  & 0.18 (0.07)  \\ \cline{2-7} 
                    & 500   & \textbf{0.27 (0.12)} & 675.0 (284.6) & 819.1 (144.6) & 0.33 (0.15)  & 0.18 (0.05)  \\ \cline{2-7} 
\hline
\multirow{3}{*}{150} & 5    & \textbf{0.43 (0.39)} & 677.0 (252.8) & 848.4 (146.4) & 0.27 (0.08)  & 0.22 (0.04)  \\ \cline{2-7} 
                    & 50    & \textbf{0.31 (0.23)} & 679.0 (254.0) & 848.0 (146.5) & 0.28 (0.09)  & 0.22 (0.07)  \\ \cline{2-7} 
                    & 500   & \textbf{0.36 (0.18)} & 649.0 (251.1) & 816.7 (163.6) & 0.28 (0.09)  & 0.22 (0.06)  \\ \cline{2-7} 
\hline
\end{tabular}}
\end{table}

\subsection{Small-Scale Physical Experiments}

To validate the efficacy of our proposed method in a real-world setting, we conduct experiments using two DAVIS346 event-based cameras, each featuring a resolution of \( 346 \times 260 \) and an image capture rate of 25 Hz. The cameras' extrinsic parameters, serving as ground truth, are calibrated using Zhang's method \cite{zhang2000flexible}. To ensure precise synchronization, the cameras are interconnected using a dedicated synchronization cable.

We fabricated a miniature drone model with dimensions of \( 3cm \times 3cm \times 1cm\) using 3D printing techniques. The model is suspended by a thin wire, allowing for free movement within the intersecting fields of view of the camera setup. To validate our simulation experiments, the angle between the cameras' lines of sight is systematically varied. Each angle configuration is rigorously confirmed using Zhang's calibration technique [23]. To introduce a controlled time offset, \( t_d \), we set \( t_d \) to predetermined values of either 5 ms, 50 ms, or 500 ms, and incorporate this offset into the timestamps of events in one of the data streams.

Utilizing our algorithm, we compute the optimal \( t_d^* \) and evaluate its deviation from the known ground truth. To quantify the algorithm's performance in estimating extrinsic parameters, we transform the matrix of estimated parameters into the rotation and translation vectors of camera \( C_2 \) relative to camera \( C_1 \). We then calculate the angular discrepancies between these estimated vectors and their ground truth counterparts. The error metrics, along with their associated standard deviations, are reported in \prettyref{tab:mini_10} and \prettyref{tab:mini_20}. Our method demonstrates high accuracy in the estimation of both the time offset \( t^*_d \) and the extrinsic parameters \( \mathbf{R} \) and \( \mathbf{t} \).

\begin{table}[t]
\centering
\caption{20-Second Small-Scale Physical Experiments}
\label{tab:mini_10}
\scalebox{0.73}{
\begin{tabular}{|c|c|c|c|c|c|}
\hline
\multirow{2}{*}{Dataset} &
\multirow{2}{*}{Angle[°]} &
\multirow{2}{*}{$t_d$[ms]} &
\multirow{2}{*}{$t_{d\ error}$[ms]} &
\multirow{2}{*}{$R_{error}$[°]} &
\multirow{2}{*}{$t_{error}$[°]}\\
    &   &   &   &   &   \\
\hline

\multirow{3}{*}{data-1}  & \multirow{3}{*}{43.14} & 5  & 0.76 (0.79) & 1.84 (0.75)  & 2.71 (2.29)  \\ \cline{3-6} 
                     &                           & 50  & 1.25 (1.1) & 2.05 (1.44)  & 2.56 (2.54)  \\ \cline{3-6} 
                     &                           & 500  & 0.71 (0.69) & 2.68 (2.45)  & 3.72 (2.9)  \\ \cline{3-6} 
\hline
\multirow{3}{*}{data-2}  & \multirow{3}{*}{65.06} & 5  & 0.41 (0.27) & 0.95 (0.45)  & 0.36 (0.31)  \\ \cline{3-6} 
                     &                           & 50  & 0.47 (0.42) & 1.01 (0.66)  & 0.47 (0.33)  \\ \cline{3-6} 
                     &                           & 500  & 0.41 (0.2) & 0.92 (0.57)  & 0.44 (0.39)  \\ \cline{3-6} 
\hline
\multirow{3}{*}{data-3}  & \multirow{3}{*}{95.18} & 5  & 1.05 (0.94) & 1.01 (0.48)  & 0.39 (0.15)  \\ \cline{3-6} 
                     &                           & 50  & 1.0 (1.33) & 0.99 (0.49)  & 0.4 (0.13)  \\ \cline{3-6} 
                     &                           & 500  & 0.79 (1.09) & 0.98 (0.32)  & 0.46 (0.22)  \\ \cline{3-6} 
\hline
\multirow{3}{*}{data-4}  & \multirow{3}{*}{131.77} & 5  & 0.96 (1.09) & 0.68 (0.24)  & 0.32 (0.12)  \\ \cline{3-6} 
                     &                           & 50  & 0.58 (0.6) & 0.69 (0.17)  & 0.32 (0.11)  \\ \cline{3-6} 
                     &                           & 500  & 0.59 (0.7) & 0.73 (0.25)  & 0.33 (0.13)  \\ \cline{3-6} 
\hline
\end{tabular}}
\end{table}

\begin{table}[t]
\centering
\caption{20-Second Small-Scale Physical Experiments}
\label{tab:mini_20}
\scalebox{0.73}{
\begin{tabular}{|c|c|c|c|c|c|}
\hline
\multirow{2}{*}{Dataset} &
\multirow{2}{*}{Angle[°]} &
\multirow{2}{*}{$t_d$[ms]} &
\multirow{2}{*}{$t_{d\ error}$[ms]} &
\multirow{2}{*}{$R_{error}$[°]} &
\multirow{2}{*}{$t_{error}$[°]}\\
    &   &   &   &   &   \\
\hline
\multirow{3}{*}{data-1}  & \multirow{3}{*}{43.14} & 5  & 0.83 (0.63) & 1.25 (0.59)  & 1.7 (1.74)  \\ \cline{3-6} 
                     &                           & 50  & 1.16 (0.92) & 1.06 (0.9)  & 2.13 (1.38)  \\ \cline{3-6} 
                     &                           & 500  & 0.58 (0.67) & 1.69 (0.77)  & 2.47 (2.66)  \\ \cline{3-6} 
\hline
\multirow{3}{*}{data-2}  & \multirow{3}{*}{65.06} & 5  & 0.43 (0.31) & 0.74 (0.49)  & 0.32 (0.18)  \\ \cline{3-6} 
                     &                           & 50  & 0.21 (0.16) & 0.69 (0.37)  & 0.47 (0.21)  \\ \cline{3-6} 
                     &                           & 500  & 0.9 (1.36) & 0.54 (0.27)  & 0.34 (0.1)  \\ \cline{3-6} 
\hline
\multirow{3}{*}{data-3}  & \multirow{3}{*}{95.18} & 5  & 0.8 (0.74) & 1.01 (0.79)  & 0.38 (0.17)  \\ \cline{3-6} 
                     &                           & 50  & 0.86 (1.37) & 0.75 (0.26)  & 0.44 (0.2)  \\ \cline{3-6} 
                     &                           & 500  & 0.81 (1.3) & 0.57 (0.21)  & 0.37 (0.1)  \\ \cline{3-6} 
\hline
\multirow{3}{*}{data-4}  & \multirow{3}{*}{131.77} & 5  & 0.8 (1.26) & 0.78 (0.24)  & 0.35 (0.19)  \\ \cline{3-6} 
                     &                           & 50  & 0.28 (0.13) & 0.71 (0.17)  & 0.32 (0.16)  \\ \cline{3-6} 
                     &                           & 500  & 1.03 (1.4) & 0.68 (0.25)  & 0.37 (0.12)  \\ \cline{3-6} 
\hline
\end{tabular}}
\end{table}

\begin{figure}[t]
    \centering\includegraphics[width=1.\linewidth]{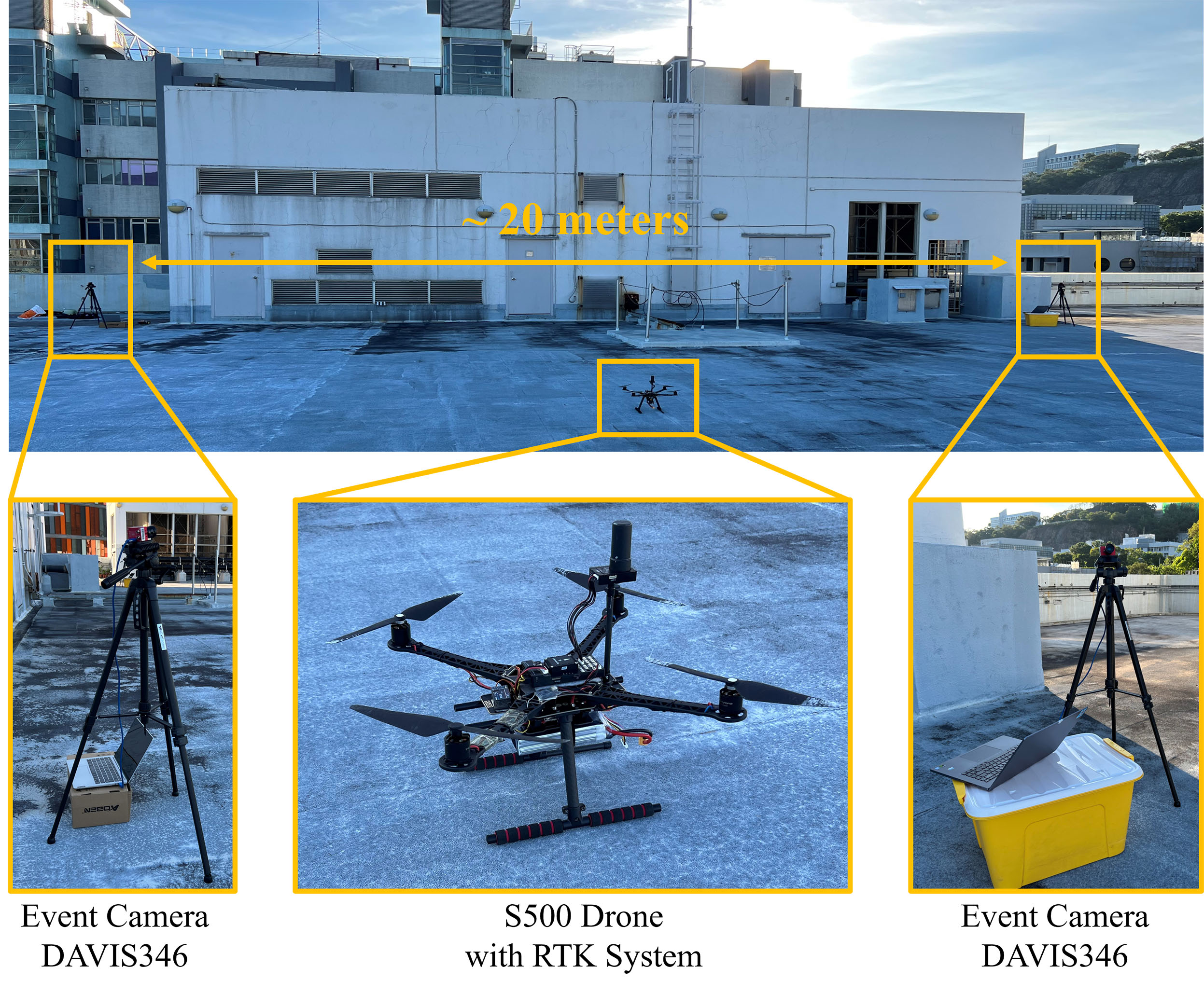}
    \caption{Experimental setup for validating drone trajectory: A drone equipped with a Real-Time Kinematic (RTK) positioning system, captured through two DAVIS346 event-based cameras}
    \label{fig:drone_setup}
\end{figure}

\subsection{Validation via Drone Trajectory in Outdoor Scenarios}

To further substantiate the robustness of our proposed methodology, we undertake an outdoor experiment featuring a drone in flight. As shown in \prettyref{fig:drone_setup}, we employ two DAVIS346 event-based cameras, each boasting a \( 346 \times 260 \) resolution. The cameras are positioned approximately 20 meters apart on the ground, with each camera connected to a dedicated laptop for data acquisition. The drone model used is a Holybro S500 equipped with a Pixhawk 4 flight controller. To obtain a ground truth trajectory, we utilize a Ublox F9P RTK system.

\begin{figure}[t]
    \centering
    \hspace{-4mm}
    \subfigure[3D Drone Trajectory]{
    \label{fig:drone_a}
    \includegraphics[width=0.251\textwidth]{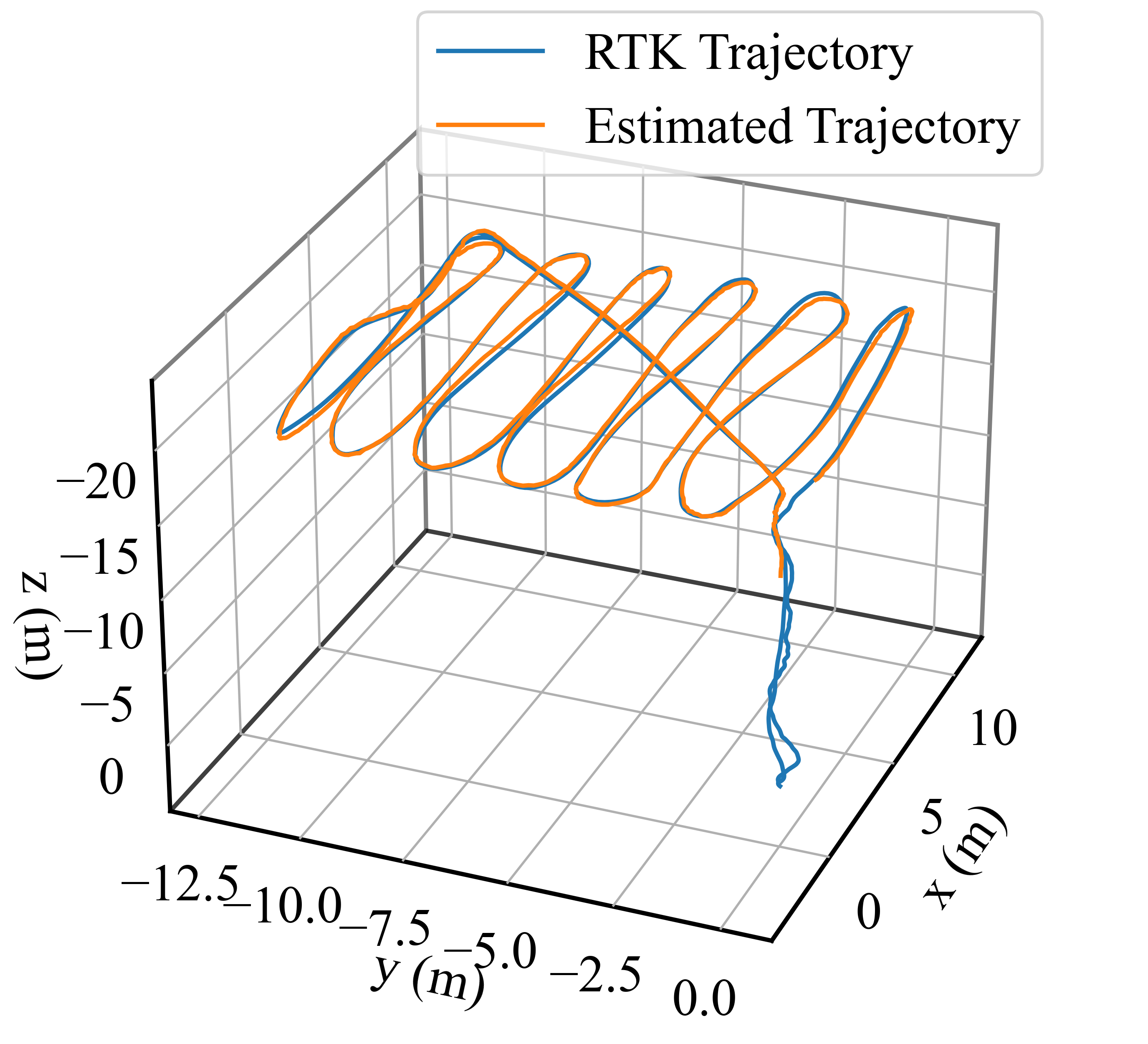}}
    \hspace{-9mm}
    \subfigure[$x$, $y$, $z$ Components]{
    \label{fig:drone_b}
    \includegraphics[width=0.274\textwidth]{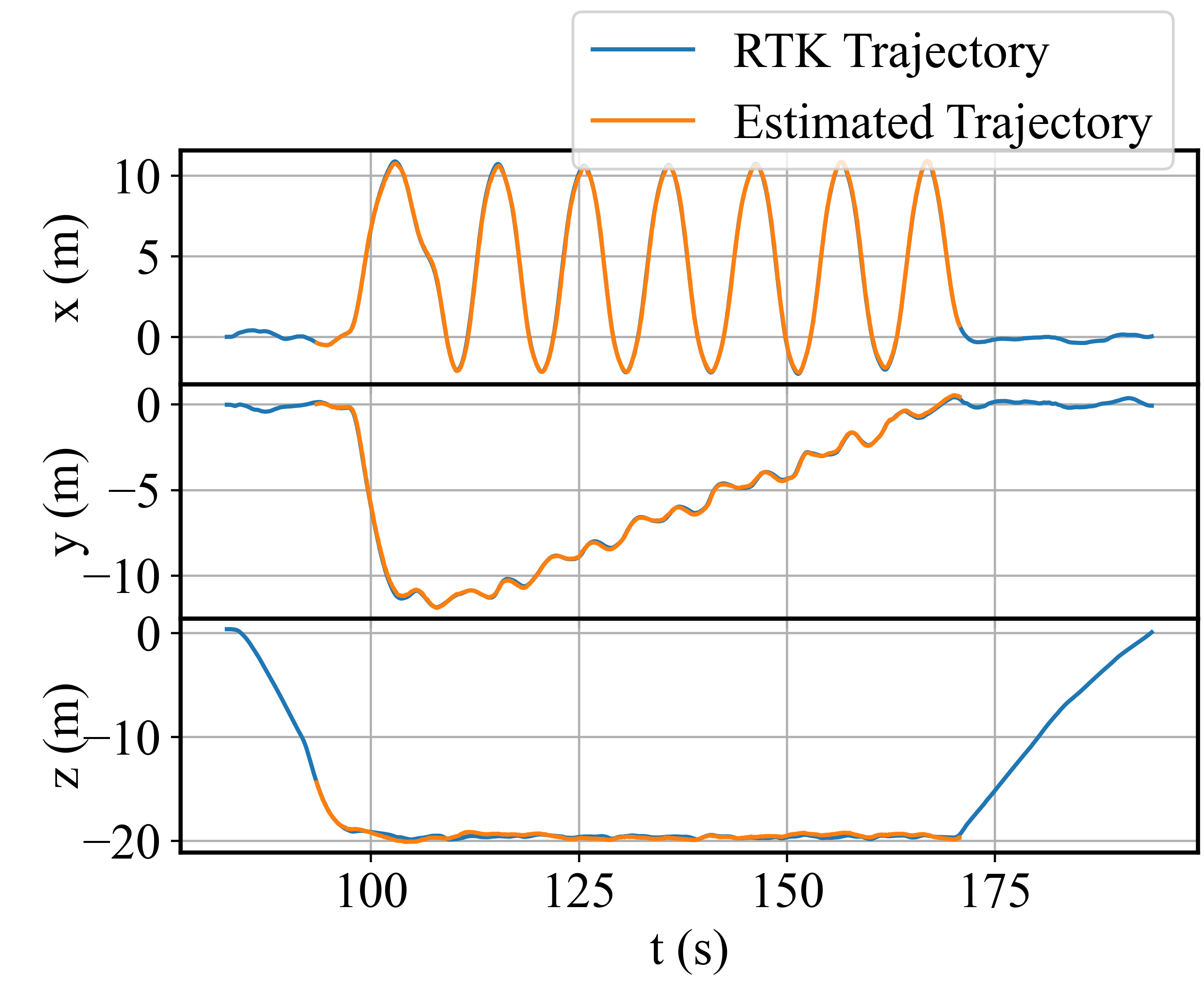}}
    \caption{(a) Comparison of the estimated drone trajectory with the ground-truth trajectory obtained via an RTK system. (b) Temporal analysis of trajectory components along the $x$, $y$, and $z$ Axes.
    }
    \label{fig:drone}
\end{figure}

We execute a 120-second flight trajectory within the cameras' common field of view. Utilizing the proposed algorithm, we process the captured event streams to calculate the optimal time offset \( t_d^* \) and estimate the cameras' extrinsic parameters. Subsequently, we reconstruct the drone's trajectory based on these calculated values. To validate our results, the reconstructed trajectory is then aligned with the RTK-recorded trajectory using the Iterative Closest Point (ICP) algorithm. The alignment is depicted in \prettyref{fig:drone_a}, with detailed translations along the \( x \), \( y \), and \( z \)-axes illustrated in \prettyref{fig:drone_b}. Our reconstructed trajectories, incorporating both estimated extrinsic parameters and event timing offsets, exhibit high accuracy.

\section{Conclusions}

In this study, we present a novel methodology for achieving millisecond-level temporal synchronization between stereo event-based cameras, eliminating the need for external mechanisms like cables. Additionally, our technique offers accurate estimations of the cameras' extrinsic parameters. Our approach will facilitate further applications in long-range stereo event-based vision.

\addtolength{\textheight}{-12cm}   








\bibliographystyle{IEEEtran}
\bibliography{IEEEabrv,references}

\end{document}